%% file: paper.tex
\documentclass{article} % For LaTeX2e
\usepackage{nips14submit_e,times}
\usepackage{hyperref}
\usepackage{url}
\usepackage{multirow}

\usepackage{framed}
\usepackage{algorithmic}
\usepackage{graphicx} 
\usepackage{subfigure}
\usepackage{amssymb}
\usepackage{amsmath}
\usepackage{amsthm}
\usepackage{caption}
\usepackage{listings}

\usepackage{verbatimbox}

\usepackage[compact]{titlesec}
\titlespacing{\section}{0pt}{0.5ex}{0.3ex}
\titlespacing{\subsection}{0pt}{0.2ex}{0ex}
\titlespacing{\subsubsection}{0pt}{0.1ex}{0ex}

\usepackage{paralist}

\makeatletter
\ifcase \@ptsize \relax% 10pt
  \newcommand{\miniscule}{\@setfontsize\miniscule{4}{5}}% \tiny: 5/6
\or% 11pt
  \newcommand{\miniscule}{\@setfontsize\miniscule{5}{6}}% \tiny: 6/7
\or% 12pt
  \newcommand{\miniscule}{\@setfontsize\miniscule{5}{6}}% \tiny: 6/7
\fi
\makeatother

\newcommand {\aplt} {\ {\raise-.5ex\hbox{$\buildrel<\over\sim$}}\ }

\newcommand{\pdata}{p_{\text{Data}}}
\newcommand{\pnoise}{p_{\text{Noise}}}

\newcommand{\eqn}[1]{Eqn.~\ref{eqn:#1}}
\newcommand{\fig}[1]{Fig.~\ref{fig:#1}}
\newcommand{\tab}[1]{Table~\ref{tab:#1}}

\newcommand {\R}{\mathbb {R}}
\def\etal{{\textit{et~al.~}}}

\title{Deep Generative Image Models using a \\Laplacian Pyramid of
  Adversarial Networks} 

\author{Emily Denton\thanks{denotes equal contribution.}\\
Dept. of Computer Science\\
Courant Institute\\
New York University
\And
{Soumith Chintala$^*$ \hspace{6mm} Arthur Szlam \hspace{6mm} Rob Fergus}\\
Facebook AI Research \\
New York
}

\nipsfinalcopy % Uncomment for camera-ready version

\begin{document}

\maketitle

\vspace{-5mm}
\begin{abstract}
  In this paper we introduce a generative parametric model capable of producing
  high quality samples of natural images. Our
  approach uses a cascade of convolutional networks within a Laplacian pyramid framework to generate
  images in a coarse-to-fine fashion. At each level of the pyramid, a separate generative convnet
  model is trained using the Generative Adversarial Nets (GAN) approach 
  \cite{Goodfellow14}. Samples
  drawn from our model are of significantly higher quality than
  alternate approaches. In a quantitative assessment by human evaluators, our
  CIFAR10 samples were mistaken for real images around 40\%  of the
  time, compared to 10\% for samples drawn from a GAN baseline model. We also show samples from
  models trained on the higher resolution images of the LSUN scene dataset.
\end{abstract}

\section{Introduction}

Building a good generative model of natural images has been a fundamental
problem within computer vision.
% If such a model existed, it would facilitate effective unsupervised learning algorithms, forming the decoder that reconstructs the input example from the latent features. 
However, images are complex and high dimensional, making them hard to model well, despite extensive efforts. 
%In this paper we present a relatively simple approach for training generative models of natural scenes and objects whose samples look signifigbetter than other approaches.
Given the difficulties of modeling entire scene at high-resolution,
most existing approaches instead generate image patches. In contrast,
in this work, we propose an approach that is able to generate plausible looking scenes at
$32 \times32$ and  $64\times64$. To do this, we exploit the 
multi-scale structure of natural images, building a series of
generative models, each of which captures image structure at a particular
scale of a Laplacian pyramid \cite{Burt83thelaplacian}. This strategy breaks the original problem into a
sequence of more manageable stages. At each scale we train a
convolutional network-based generative model using the Generative Adversarial
Networks (GAN) approach of Goodfellow \etal \cite{Goodfellow14}.  Samples
are drawn in a coarse-to-fine fashion, commencing with a low-frequency
residual image. The second stage samples the band-pass structure at
the next level, conditioned on the sampled residual. Subsequent levels
continue this process, always conditioning on the output from the
previous scale, until the final level is reached. Thus drawing samples
is an efficient and straightforward procedure: taking random vectors
as input and running forward through a cascade of deep convolutional
networks (convnets) to produce an image.

Deep learning approaches have proven highly effective at
discriminative tasks in vision, such as object
classification \cite{deng2009imagenet}. However, the same level of success has
not been obtained for generative tasks, despite numerous efforts
\cite{hinton2006reducing,salakhutdinov2009deep,Vincent08}. Against this background, our
proposed approach makes a significant advance in that it is
straightforward to train and sample from, with the resulting
samples showing a surprising level of visual fidelity, indicating a better density model than prior methods.

% This basic model can be utilized in a number of ways: (i) if labeled training
% data is available, the generative process can also be conditioned on
% object class; (ii) by starting the sampling cascade with an existing
% real image allows it to be super-resolved; (iii) in-painting can be
% performed by downsampling the image to a coarse resolution, using
% standard methods to in-paint the whole, then run image through the
% cascade.

\subsection{Related Work}

Generative image models are well studied, falling into two main
approaches: non-parametric and parametric. The former copy patches from
training images to perform, for example, texture
synthesis \cite{efros99} or super-resolution
\cite{freeman2002example}. More ambitiously, entire portions of an
image can be in-painted, given a sufficiently large training dataset
\cite{hays2007scene}. Early parametric models addressed the easier problem of texture
synthesis \cite{de1997multiresolution,zhu1998filters,portilla2000},
with Portilla \& Simoncelli \cite{portilla2000} making use of a
steerable pyramid wavelet representation
\cite{simoncelli1992shiftable}, similar to our use of a Laplacian pyramid.  For image processing tasks, models
based on marginal distributions of image gradients are effective
\cite{olshausenfield,Roth05fieldsof}, but are only designed for image restoration
rather than being true density models (so cannot sample an actual
image). Very large Gaussian mixture models
\cite{ZoranWiess} and sparse coding models of image patches \cite{wright2010sparse} can also
be used but suffer the same problem.

A wide variety of deep learning approaches involve generative
parametric models. Restricted Boltzmann machines
\cite{hinton2006reducing,krizhevsky2010factored,Osindero08,ranzato2013modeling}, Deep Boltzmann
machines \cite{salakhutdinov2009deep,eslami2014shape}, Denoising
auto-encoders \cite{Vincent08} all have a generative decoder that
reconstructs the image from the latent representation. Variational
auto-encoders \cite{Kingma14,Rezende14} provide probabilistic
interpretation which facilitates sampling. However, for all these methods
convincing samples have only been shown on simple datasets such as
MNIST and NORB, possibly due to 
training complexities which limit their applicability to larger and more realistic images.

Several recent papers have proposed novel generative models.
Dosovitskiy \etal \cite{dosovitskiy2014} showed how a convnet can draw
chairs with different shapes and viewpoints. While our model also
makes use of convnets, it is able to sample general scenes and
objects. The DRAW model of Gregor
\etal \cite{GregorDGW15} used an attentional mechanism with an RNN to generate images via a
trajectory of patches, showing samples of MNIST and CIFAR10
images. Sohl-Dickstein \etal \cite{Sohl-Dickstein15} use a
diffusion-based process for deep unsupervised learning and the
resulting model is able to produce reasonable CIFAR10 samples. Theis
and Bethge \cite{Theis2015c} employ LSTMs to capture spatial
dependencies and show convincing inpainting results of natural textures.

Our work builds on the GAN approach of Goodfellow \etal
\cite{Goodfellow14} which works well for smaller images (e.g.~MNIST) but cannot
directly handle large ones, unlike our method. Most relevant to our
approach is the preliminary work of Mirza and Osindero \cite{MirzaO14}  and Gauthier \cite{Gauthier14} who both propose
conditional versions of the GAN model. The former shows MNIST samples,
while the latter focuses solely on frontal face images. Our approach
also uses several forms of conditional GAN model but is much more
ambitious in its scope.

\input{approach}

\input{experiments}

\section{Discussion}
By modifying the approach in \cite{Goodfellow14} to better respect the structure of images, 
we have proposed a conceptually simple generative model that is able
to produce high-quality sample images that
are both qualitatively and quantitatively better than other deep
generative modeling approaches. 
A key point in our work is  giving up any ``global'' notion of fidelity, and instead breaking the generation into plausible successive refinements. 
We note that many other signal modalities have a multiscale structure that may benefit from a similar approach. 

\vspace{25mm}

%At higher resolutions, the samples remain sharp but do not develop further object structure, perhaps due to the independent training of each scale, which allows ``drift'' of the samples away
%from plausible images. This could remedied by conditioning on the
%output of previous scales. 
%We also notice that for our datasets the $8\rightarrow16$ scale transition is a critical one where much of the object structure develops. Improving the fidelity of models at this scale seems a priority for improving sampling quality. 

\section*{Appendix A}
To describe the log-likelihood computation in our model, let us
consider a two scale pyramid for the moment.  Given a (vectorized) $j\times j$ image $I$, denote by $l=d(I)$ the coarsened image, and $h=I-u(d(I))$ to be the high pass. 
In this section, to simplify the computations, we use a slightly different $u$ operator than the one used to generate the images displayed in \fig{cifar_samples}.  Namely, here we take 
$d(I)$ to be the mean over each disjoint block of $2\times 2$ pixels,
and take $u$ to be the operator that removes the mean from each
$2\times 2$ block.   Since $u$ has rank $3d^2/4$, in this section, we
write $h$ in an orthonormal basis of the range of $u$, then the
(linear) mapping from $I$ to $(l,h)$ is unitary. We now build a probability density $p$ on $\R^{d^2}$ by 
\vspace{-1mm}
\[p(I) = q_0(l,h) q_1(l)= q_0(d(I),h(I)) q_1(d(I));\] 
\vspace{-1mm}
 in a moment we will carefully define the functions $q_i$.   For now, suppose that $q_i\geq 0$, $\int q_1(l) \,dl =1$, and for each fixed $l$, $\int q_0(l,h) \,dh =1$.
Then we can check that $p$ has unit integral:
\vspace{-2mm}
\[\int p \,dI = \int q_0(d(I),h(I)) q_1(d(I))  dI =  \int \int q_0(l,h) q_1(l) \, dl \, dh = 1.\]
%\vspace{-1mm}
%This means that 
%\[\int p(x) \left|\text{det}\, A\right| = \int q(Ax)\left|\text{det}\, A\right| = 1\]
%and so  
%\[\int p(x) = 1/\left|\text{det}\, A\right|\]
Now we define the $q_i$ with Parzen window approximations to the densities of each of the scales.  For $q_1$, we take a set of training samples 
$l_1,....,l_{N_0}$, and construct the density function $q_1(l) \sim
\sum_{i=1}^{N_1} e^{||l-l_i||^2/\sigma_1}$. 
%\[q_1(l) \sim \sum_{i=1}^{N_1} e^{||l-l_i||^2/\sigma_1}.\]
We fix $l=d(I)$ to define $q_0(I) = q_0(l,h) \sim \sum_{i=1}^{N_0} e^{||h-h_i||^2/\sigma_0}$.%\footnote{Note that when defined this way, it is not obvious that $q_1$ is a measurable function, as the choice of $h_i$ by the upsampling model is different for every $l$ (and in fact depends on the random seed we used to sample).  However, because the mapping from fixed ``noise variable'' and coarse image to refinement is the forward of a convolutional net, and so is continuous, if we use the same random seeds for each $I$,  $q_1$ is measurable.}. Using this fixed $l$, we generate $N_0$ points $h_1,...,h_{N_1}$ from the refinement model, and define
%\[q_0(I)= q_0(l,h) \sim \sum_{i=1}^{N_0} e^{||h-h_i||^2/\sigma_0}.\]
For pyramids with more levels, we continue in the same way for each of the finer scales.  Note we
always use the true low pass at each scale, and measure the true high
pass against the high pass samples generated from the model. Thus for a
pyramid with $K$ levels, the final
log likelihood will be: $\log(q_K(l_K)) + \sum_{k=0}^{K-1}
\log(q_k(l_k,h_k))$.

\clearpage
%\scriptsize
\small
\bibliography{bibliography} 
\bibliographystyle{ieee}

\end{document}

%% file: approach.tex
\section{Approach}
The basic building block of our approach is the generative adversarial
network (GAN) of Goodfellow \etal \cite{Goodfellow14}. After reviewing
this, we introduce our LAPGAN model
which integrates a conditional form of GAN model into the framework of a Laplacian pyramid.

\subsection{Generative Adversarial Networks}
The GAN approach \cite{Goodfellow14} is a framework for training
generative models, which we briefly explain in the context of image data.
The method pits two networks against one another: a generative model
$G$ that captures the data distribution and a discriminative model $D$ that distinguishes between samples drawn from $G$ and images drawn from the training data. 
In our approach, both $G$ and $D$ are convolutional networks. The former takes as input a noise vector $z$ drawn from a
distribution $\pnoise(\mathbf z)$ and outputs an image $\tilde{h}$. The
discriminative network $D$ takes an image as input stochastically
chosen (with equal probability) to be either $\tilde{h}$ -- as
generated from $G$, or $h$ -- a real image drawn from the training
data $\pdata(\mathbf h)$. $D$ outputs a scalar probability, which is
trained to be high if the input was real and low if generated from $G$. 
A minimax objective is used to train both models together:
\vspace{-1mm}
\begin{equation}
\vspace{-0mm}
\label{eqn:gan}
\min_G \max_D \mathbb{E}_{h\sim \pdata(\mathbf h)} [\log D(h)]  +
\mathbb{E}_{z \sim \pnoise(\mathbf z)} [\log(1 - D(G(z)))]
\vspace{-0mm}
\end{equation}
This encourages $G$ to fit $\pdata(\mathbf h)$ so as to fool $D$ with its
generated samples $\tilde{h}$. Both $G$ and $D$ are trained by backpropagating
the loss in \eqn{gan} through their respective models to update the parameters.

The conditional generative adversarial net (CGAN) is an extension of
 the GAN where both networks $G$ and $D$ receive an additional
 vector of information $l$ as input. This might contain, say,
 information about the class of the training example $h$. The loss
 function thus becomes
\begin{equation}
\label{eqn:cgan}
\min_G \max_D \mathbb{E}_{h,l\sim \pdata(\mathbf h,\mathbf l)} [\log
D(h,l)]  + \mathbb{E}_{z \sim \pnoise(\mathbf z), l \sim p_l(\mathbf l)} [\log(1 - D(G(z,l),l))]
\end{equation}
where $p_l(\mathbf l)$ is, for example, the prior distribution over
classes. This model allows the output of the generative model
to be controlled by the conditioning variable $l$. Mirza and Osindero
\cite{MirzaO14}  and Gauthier \cite{Gauthier14} both explore this
model with experiments on MNIST and faces, using $l$ as a class
indicator. In our approach, $l$ will be another image, generated from
another CGAN model.

\subsection{Laplacian Pyramid} 
The Laplacian pyramid \cite{Burt83thelaplacian} is a linear invertible image
representation consisting of a set of band-pass images, spaced an
octave apart, plus a low-frequency residual. Formally, let $d(.)$ be a
downsampling operation which blurs and decimates a $j \times j$ image $I$, so that
$d(I)$ is a new image of size $j/2 \times j/2$. Also, let $u(.)$ be an
upsampling operator which smooths and expands $I$ to be twice the
size, so $u(I)$ is a new image of size $2j \times 2j$. 
We first build a Gaussian pyramid $\mathcal{G}(I) =
[I_0,I_1,\ldots,I_K]$, where $I_0=I$ and $I_k$ is $k$ repeated
  applications\footnote{i.e.~$I_2=d(d(I))$.} of $d(.)$ to $I$. $K$ is the number of levels in the
pyramid, selected so that the final level has very small spatial extent ($\leq
8\times8$ pixels). 

The coefficients $h_k$ at each level $k$ of the Laplacian pyramid
$\mathcal{L}(I)$ are constructed
by taking the difference between adjacent levels in the Gaussian
pyramid, upsampling the smaller one with $u(.)$ so that the sizes are
compatible: 
\vspace{-2mm}
\begin{equation}
h_k = \mathcal{L}_k(I) = \mathcal{G}_k(I) - u(\mathcal{G}_{k+1}(I))
= I_k - u(I_{k+1}) 
\vspace{-2mm}
\label{eqn:lapbuild}
\end{equation} 
Intuitively, each level captures
image structure present at a particular scale. The final level of the Laplacian pyramid $h_K$
is not a difference image, but a low-frequency residual equal to the final Gaussian pyramid
level, i.e.~ $h_K=I_K$. Reconstruction from a Laplacian pyramid coefficients $[h_1,\ldots,h_K]$ is
performed using the backward recurrence:
\vspace{-2mm}
\begin{equation}
I_k=u(I_{k+1}) + h_k 
\vspace{-2mm}
\label{eqn:laprecon}
\end{equation} 
which is started with $I_K=h_K$ and the reconstructed image being
$I = I_o$. In other words, starting at the
coarsest level, we repeatedly upsample and add the difference image $h$ at the next
finer level until we get back to the full resolution image.

\subsection{Laplacian Generative Adversarial Networks (LAPGAN)}

\begin{figure}[b!]
\vspace{-3mm}
\centering
        \includegraphics[width=\textwidth]{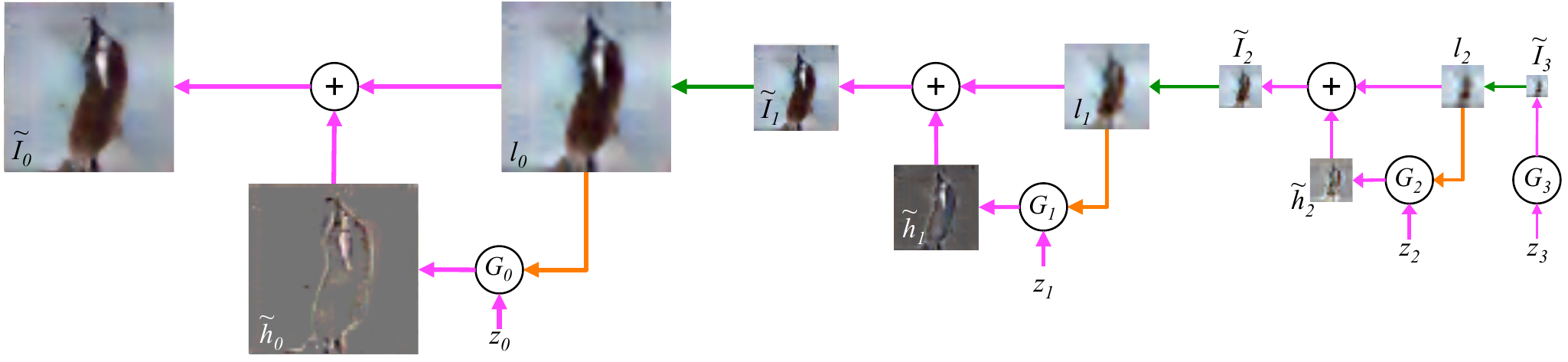}
\vspace{-3mm}
\caption{\small The sampling procedure for our LAPGAN model. We start
  with a noise sample $z_3$ (right side) and use a generative model $G_3$ to generate
  $\tilde{I}_3$. This is upsampled (green arrow) and then used as the
  conditioning variable (orange arrow) $l_2$ for the generative model at the next
  level, $G_2$. Together with another noise sample $z_2$, $G_2$
  generates a difference image $\tilde{h}_2$ which is added to $l_2$
  to create $\tilde{I}_2$. This process repeats across two subsequent
  levels to yield a final full resolution sample $I_0$.}
\label{fig:sampling}
\vspace{-3mm}
\end{figure}

 Our proposed approach combines the conditional GAN model with a
Laplacian pyramid representation. The model is best explained by
first considering the sampling procedure. Following training (explained below), we have a set of
generative convnet models $\{G_0,\ldots,G_K\}$, each of which captures the
distribution of coefficients $h_k$ for natural images at a different level of the Laplacian
pyramid. Sampling an image is akin to the reconstruction procedure in
\eqn{laprecon}, except that the generative models are used to produce the $h_k$'s:
%\vspace{-2mm}
\begin{equation}
\tilde{I}_k=u(\tilde{I}_{k+1}) + \tilde{h}_k = u(\tilde{I}_{k+1}) + G_k(z_k,u(\tilde{I}_{k+1})) 
%\vspace{-2mm}
\label{eqn:sampbuild}
\end{equation} 
The recurrence starts by setting $\tilde{I}_{K+1}=0$ and using the model at the final level $G_K$ to generate a residual
image $\tilde{I}_K$ using noise vector $z_K$: $\tilde{I}_K =
G_K(z_K)$. Note that models at all levels except the final
are conditional generative models that take an upsampled version of
the current image $\tilde{I}_{k+1}$ as a conditioning
  variable, in addition to the noise vector $z_k$. \fig{sampling} shows this procedure in action for a
  pyramid with $K=3$ using 4 generative models to sample a
  $64\times64$ image.   

\begin{figure}
\vspace{-2mm}
\centering
        \includegraphics[width=\textwidth]{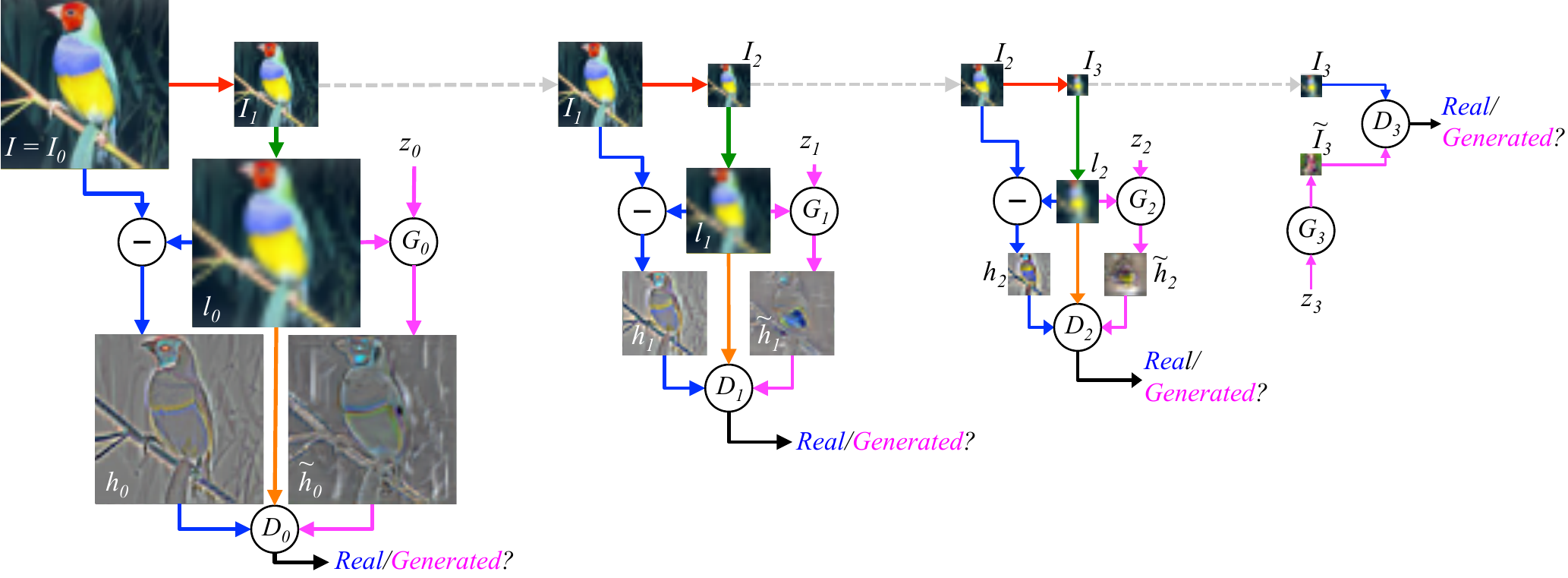}
\vspace{-4mm}
\caption{\small The training procedure for
  our LAPGAN model. Starting with
  a 64x64 input image $I$ from our training set (top left): (i) we take
  $I_0=I$ and blur and downsample it by a factor of two (red arrow) to
  produce $I_1$; (ii) we upsample $I_1$ by a factor of two (green
  arrow), giving a low-pass version $l_0$ of $I_0$;
  (iii) with equal probability we use $l_0$ to create {\em either}
  a real {\em or} a generated example for the discriminative model $D_0$. In the real
  case (blue arrows), we compute high-pass $h_0=I_0-l_0$ which is input to $D_0$ that computes the
  probability of it being real vs generated. In the generated
  case (magenta arrows), the generative network $G_0$ receives as
  input a random noise vector $z_0$ and $l_0$. It outputs a generated high-pass
  image $\tilde{h}_0=G_0(z_0,l_0)$, which is input to $D_0$. In both the real/generated cases, $D_0$
 also receives $l_0$ (orange arrow). Optimizing \eqn{cgan}, $G_0$
 thus learns to generate
  realistic high-frequency structure $\tilde{h}_0$ consistent with the low-pass
  image $l_0$. The same procedure is repeated at scales 1 and 2,
  using $I_1$ and $I_2$. Note that the
  models at each level are trained independently. At level 3, $I_3$
  is an 8$\times$8 image, simple enough to be modeled directly with a
  standard GANs $G_3$ \& $D_3$. }
\label{fig:training}
\vspace{-5mm}
\end{figure}

The generative models $\{G_0,\ldots,G_K\}$ are trained using the CGAN
approach at each level of the pyramid. Specifically, we construct a
Laplacian pyramid from each training image $I$. At each level we make
a stochastic choice (with equal probability) to either (i) construct the coefficients $h_k$ either using the standard procedure from
\eqn{lapbuild}, or (ii) generate them using $G_k$:
%\vspace{-2mm}
\begin{equation}
\tilde{h}_k=G_k(z_k,u(I_{k+1}))  
%\vspace{-2mm}
\label{eqn:samprecon}
\end{equation} 
Note that $G_k$ is a convnet which uses a coarse scale
version of the image $l_k=u(I_{k+1})$ as an input, as well as noise
vector $z_k$. $D_k$ takes as input $h_k$ or
$\tilde{h}_k$, along with the low-pass image
$l_k$ (which is explicitly added to $h_k$ or
$\tilde{h}_k$ before the first
convolution layer), and predicts if the image was real or generated. At the final scale of the pyramid, the low
frequency residual is sufficiently small that it can be directly modeled
with a standard GAN: $\tilde{h}_K=G_K(z_K)$ and $D_K$ only has $h_K$
or $\tilde{h}_K$ as input. The framework is
illustrated in \fig{training}.

Breaking the generation into successive refinements is the key idea in
this work.  Note that we give up any ``global'' notion of fidelity; we
never make any attempt to train a network to discriminate between the
output of a cascade and a real image and instead focus on making
each step plausible. Furthermore, the independent training of each pyramid
level has the advantage that it is far more difficult for the model
to memorize training examples -- a hazard when high capacity deep
networks are used.

% By using the multi-scale structure of the pyramid, the complex
% disitribution of natural images is broken down into manageable pieces,
% each of which can be independently modeled by a deep convnet
% model. This avoids having to directly model an entire image at
% full-resolution in one go and, as our experiments show, is crucial to
% obtaining plausible samples. 
As described, our model is trained in an unsupervised manner. However,
we also explore variants that utilize class labels. This is done by
add a 1-hot vector $c$, indicating class identity, as another
conditioning variable for $G_k$ and $D_k$. 

\section{Model Architecture \& Training}
We apply our approach to three datasets: (i) {\bf CIFAR10} --
32$\times$32 pixel color images of 10 different classes, 100k training
samples with tight crops of objects; (ii) {\bf STL} --  96$\times$96
pixel color images of 10 different classes, 100k training samples (we
use the unlabeled portion of data); and (iii) {\bf
  LSUN} \cite{Xiao15} --
$\sim$10M images of 10 different natural scene types, downsampled to
64$\times$64 pixels.% \rob{Check}.

For each dataset, we explored a variety of architectures
for $\{G_k,D_k\}$. We now detail the best performing models, selected
using a combination of log-likelihood and visual appearance of the
samples. Complete {\tt Torch} specification files for all models are provided in 
supplementary material \cite{supp}. For all models, the noise vector $z_k$ is drawn from a uniform [-1,1] distribution.

\subsection{CIFAR10 and STL}
\vspace{-2mm}
\noindent {\bf Initial scale:} This operates at $8\times8$ resolution,
using densely connected nets for both $G_K$ \& $D_K$ with 2 hidden layers
and ReLU non-linearities. $D_K$ uses Dropout and has 600 units/layer
vs 1200 for $G_K$. $z_K$ is a 100-d vector.
%, each element drawn from a
%uniform $[-1,1]$ distribution.

\noindent {\bf Subsequent scales:} For CIFAR10, we boost the training
set size by taking four $28\times28$ crops from the original
images. Thus the two subsequent levels of the pyramid are
$8\rightarrow14$ and
$14\rightarrow28$. For STL, we have 4 levels going
from $8\rightarrow16\rightarrow32\rightarrow64\rightarrow96$.  
For both datasets, $G_k$ \& $D_k$ are convnets with 3 and 2 layers,
respectively (see \cite{supp}). 
%For CIFAR, the $8\rightarrow14$ and $14\rightarrow28$ levels use 64
%and 128 feature maps/layer (with ReLUs) respectively. 
%For STL10 \rob{TODO}.
The noise input $z_k$ to $G_k$ is presented as a 4th ``color plane''
to low-pass $l_k$, hence its dimensionality varies with the pyramid
level. 
%Again, each element is drawn from a $[-1,1]$ uniform
%distribution. This form of $z_k$ is used for all other datasets as well. 
For CIFAR10, we also explore a class conditional version of the model,
where a vector $c$ encodes the label. This is integrated into
$G_k$ \& $D_k$ by passing it through a linear layer whose output is reshaped
into a single plane feature map which is then concatenated with the 1st
layer maps.
The loss in \eqn{cgan} is trained using SGD with an initial learning rate
of 0.02, decreased by a factor of $(1+4\times10^{-5})$ at each
epoch. Momentum starts at 0.5, increasing by 0.0008 at epoch up to a
maximum of 0.8. During training, we monitor log-likelihood using a
Parzen-window estimator and retain the best performing model. 
%CIFAR training takes a few hours, while STL10 takes several days. 
Training time depends on the models size and pyramid level, with smaller models taking hours to train and larger models taking several days. 

\subsection{LSUN}
\vspace{-2mm}
%\noindent {\bf Initial scale:} This operates at $8\times8$ resolution,
%using densely connected nets for both $G_K/D_K$ with 2 hidden layers
%and ReLU non-linearities. $D_K$ uses Dropout and has 600 units/layer
%vs 1200 for $G_K$. $z_K$ is a 100-d vector, each element drawn from a
%uniform $[-1,1]$ distribution. \rob{Duplicate}
The larger size of this dataset allows us to train a separate LAPGAN model for each the 10 different scene
classes. During evaluation, so that we may understand the variation captured by
our models, we commence the sampling process with validation set
images\footnote{These were not used in any way during training.}, downsampled to
$4\times4$ resolution.% \rob{Something about 4 by 4 generations?}
% (rather than using a GAN at the inital
%scale, although it makes no difference to the quality of samples).

The four subsequent scales
$4\rightarrow8\rightarrow16\rightarrow32\rightarrow64$ use a common
architecture for $G_k$ \& $D_k$ at each level. $G_k$ is a 5-layer convnet with $\{64,368,128,224\}$ feature maps and a
linear output layer. $7\times7$ filters, ReLUs,
batch normalization \cite{Ioffe15} and Dropout are used at each hidden
layer. $D_k$ has 3 hidden layers  with $\{48,448,416\}$ maps plus a sigmoid
output. See \cite{supp} for full details. Note that $G_k$ and $D_k$
are substantially larger than those used for CIFAR10 and STL, as afforded by the
larger training set.

%% file: experiments.tex
\section{Experiments}
We evaluate our approach using 3
different methods: (i) computation of log-likelihood on a held out
image set; (ii) drawing sample images from the model and (iii) a human
subject experiment that compares (a) our samples, (b) those of baseline
methods and (c) real images. 
%As a baseline, we reimplemented the standard GAN model of
%Goodfellow \etal \cite{Goodfellow14} and compared it to our LAPGAN model.

\subsection{Evaluation of Log-Likelihood}
A traditional method for evaluating generative models is to measure
their log-likelihood on a held out set of images. 
%But, like the original GAN method \cite{Goodfellow14}, our approach does
%not have a direct way of computing the probability of an image so we resort to a Parzen-window estimator based
%on samples from the model. However, the performance computation of log-likelihood
%with our multi-scale approach requires some care, the procedure being
%described in Appendix A. 
But, like the original GAN method \cite{Goodfellow14}, our approach does
not have a direct way of computing the probability of an image. 
Goodfellow \etal \cite{Goodfellow14} propose using a Gaussian Parzen window estimate to compute log-likelihoods.
Despite showing poor performance in high dimensional spaces, this
approach is the best one available for estimating likelihoods of
models lacking an explicitly represented density function. 

Our LAPGAN model allows for an alternative method of estimating
log-likelihood that exploits the multi-scale structure of the model.
This new approach uses a Gaussian Parzen window estimate to compute a
probability at each scale of the Laplacian pyramid.  We use this
procedure, described in detail in Appendix A, to compute the
log-likelihoods for CIFAR10 and STL images (both at $32\times32$
resolution). The parameter
$\sigma$ (controlling the Parzen window size) was chosen using the
validation set. 
%For this computation, we use $N_0=10k, N_1=600$ and
%select optimal values of $\sigma_0,\sigma_1$ using the validation set.
We also compute the Parzen window based log-likelihood estimates of the
standard GAN \cite{Goodfellow14} model, using 50k samples for both the
CIFAR10 and STL estimates. \tab{llh} shows our model achieving a significantly
higher log-likelihood on both datasets.  Comparisons to further approaches,
notably \cite{Sohl-Dickstein15}, are problematic due to different
normalizations used on the data.  

\begin{table}[h!]
\vspace{-3mm}
\centering
\begin{tabular}{c|c|c} 
Model & CIFAR10 & STL (@32$\times$32)\\
\hline 
\hline
GAN \cite{Goodfellow14}& -3617 $\pm$ 353& -3661 $\pm$ 347\\
LAPGAN & -1799 $\pm$ 826 & -2906 $\pm$ 728 \\
%Training set & -3561 $\pm$ 355 &  -3230 $\pm$ 438 
\end{tabular}
\vspace{-3mm}
\caption{Parzen window based log-likelihood estimates for a standard
  GAN, our proposed LAPGAN model on CIFAR10 and STL datasets.}
\label{tab:llh}
\end{table}

\subsection{Model Samples}
We show samples from models trained on CIFAR10, STL and LSUN datasets. 
Additional samples can be found in the supplementary material \cite{supp}.

\fig{cifar_samples} shows samples from our models trained on CIFAR10.
Samples from the class conditional LAPGAN are organized by class.  Our
reimplementation of the standard GAN model \cite{Goodfellow14}
produces slightly sharper images than those shown in the original
paper.  We attribute this improvement to the introduction of data
augmentation.  The LAPGAN samples improve upon the standard GAN
samples. They appear more object-like and have more clearly defined
edges.  Conditioning on a class label improves the generations as
evidenced by the clear object structure in the conditional LAPGAN
samples. The quality of these samples compares favorably with those
from the DRAW model of Gregor \etal \cite{GregorDGW15} and also
Sohl-Dickstein \etal \cite{Sohl-Dickstein15}.  The rightmost column of
each image shows the nearest training example to the neighboring
sample (in L2 pixel-space). This demonstrates that our model is not
simply copying the input examples.

\begin{figure}[h!]
\centering
\mbox{
%\hspace{5mm}
\subfigure[][]{
  \includegraphics[width=0.48\linewidth]{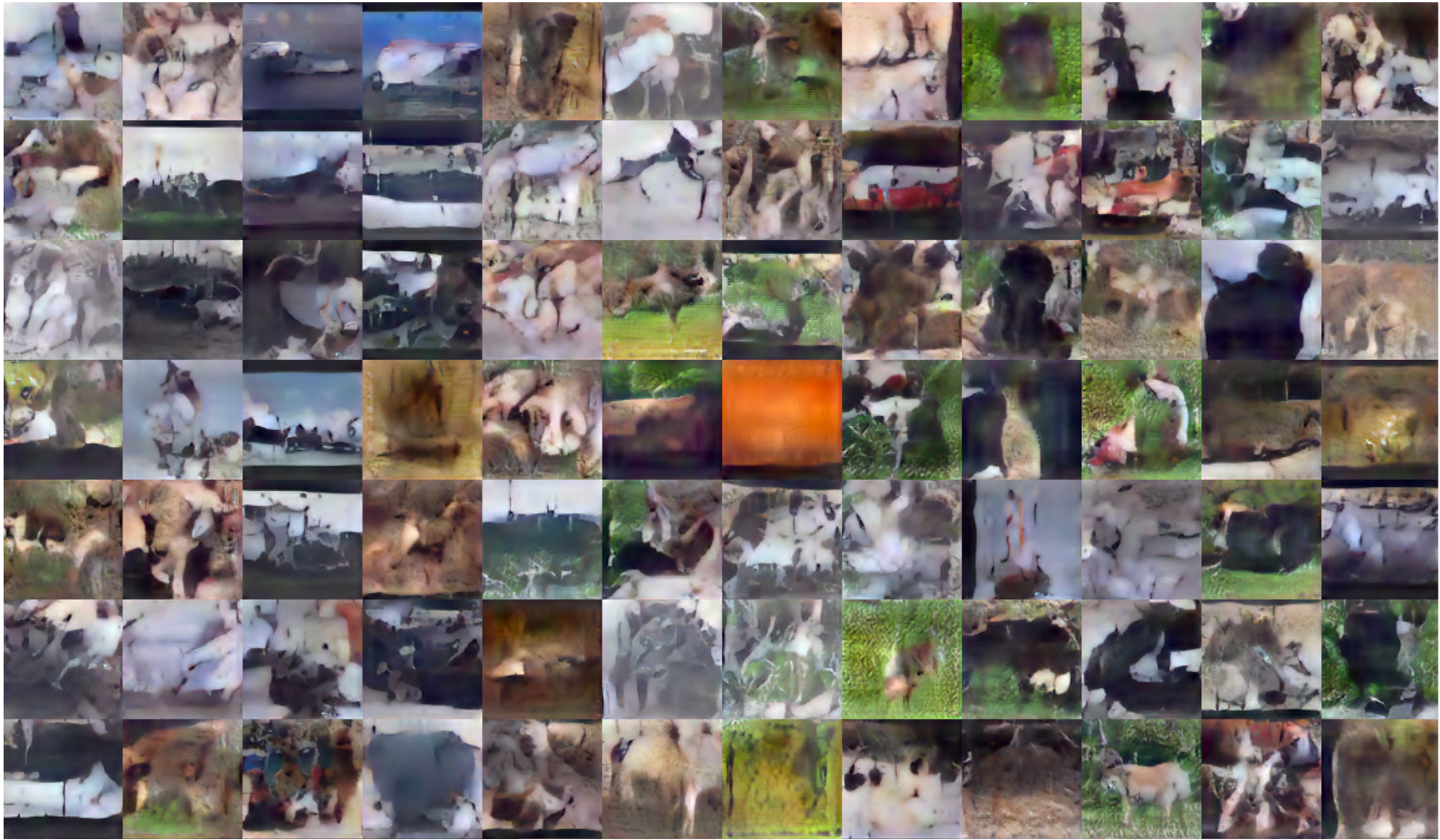} 
  \label{fig:stl_samples}
}
\hspace{1mm}
\subfigure[][]{
  \includegraphics[width=0.35\linewidth]{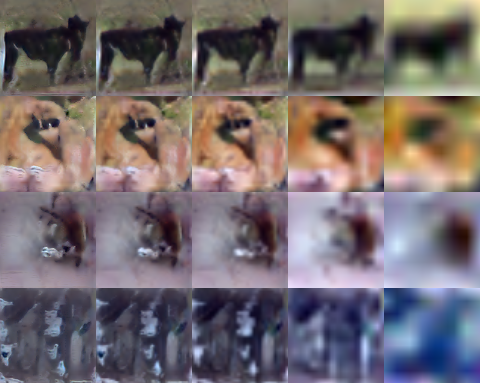} 
  \label{fig:stl_progression}
}
}
\vspace{-5mm}
\caption{ STL samples: {\bf(a)}  Random 96x96 samples from our LAPGAN model. {\bf (b)} Coarse-to-fine generation chain. }
\label{samples}
\end{figure}

\fig{stl_samples} shows samples from our LAPGAN model trained on
STL.  Here, we lose clear object shape but the samples remain
sharp.  \fig{stl_progression} shows the generation chain for random
STL samples. 

\begin{figure}[h!]
%\vspace{-4mm}
\centering
  \includegraphics[width=0.90\linewidth]{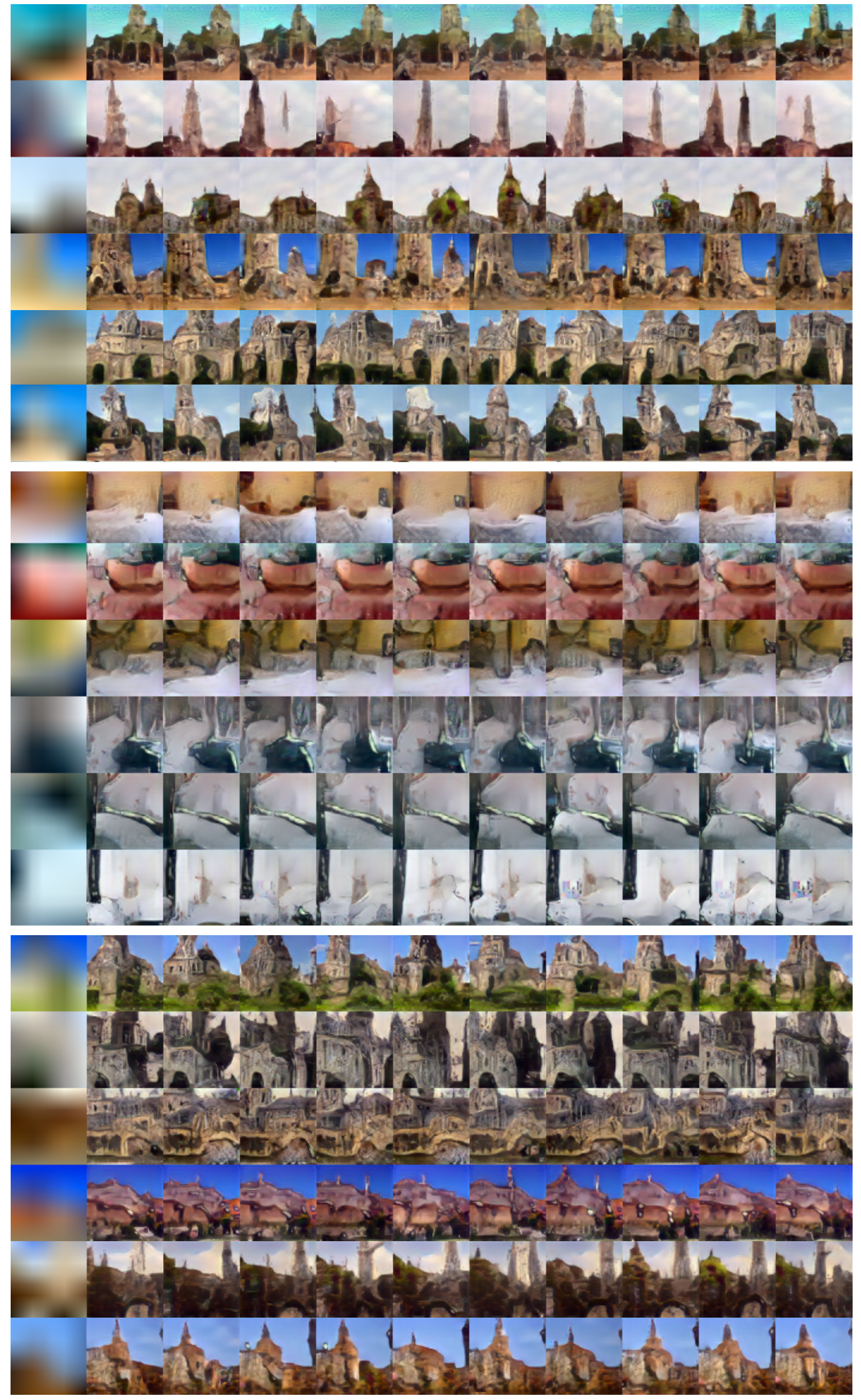}
 \vspace{-2mm}
\caption{$64\times64$ samples from three different LSUN LAPGAN models (top:
  tower, middle: bedroom, bottom: church front). The first column shows
  the $4\times4$ validation set image used to start the generation
  process, with subsequent columns showing different draws from the model. }
\label{fig:lsun_samples}
\end{figure}

\fig{lsun_samples} shows samples from LAPGAN models trained on three LSUN
categories (tower, bedroom, church front). The $4\times4$ validation image used to start the
generation process is shown in the first column, along with 10
different $64\times64$ samples, which illustrate the inherent
variation captured by the model. Collectively, these show the
models capturing long-range structure within the scenes, being able to
recompose scene elements into credible looking images. To the best of our knowledge, no other
generative model has been able to produce samples of this
complexity. The substantial gain in quality over the CIFAR10 and STL
samples is likely due to the much larger training LSUN training set
which allowed us to train bigger and deeper models.

%\fig{samples} shows samples from models trained on each of the four
%datasets. SOME DISCUSSISON OF SAMPLES. Comparison to nearest
%neighbors. We also show samples here the low-resolution image is held
%constant and only the fine scales are resampled. Also show
%trajectories between different noise vectors $z$. 
%%%%%%%%%%%%%%%%%%%%%%%%%%%%%%%%%%%%%%%%%%%%%%%%%%%%%%%%%
%% CIFAR10 samples - regular and class conditional
%%%%%%%%%%%%%%%%%%%%%%%%%%%%%%%%%%%%%%%%%%%%%%%%%%%%%%%%%
\begin{figure}[h!]
%\vspace{-4mm}
\centering
  \includegraphics[width=0.97\linewidth]{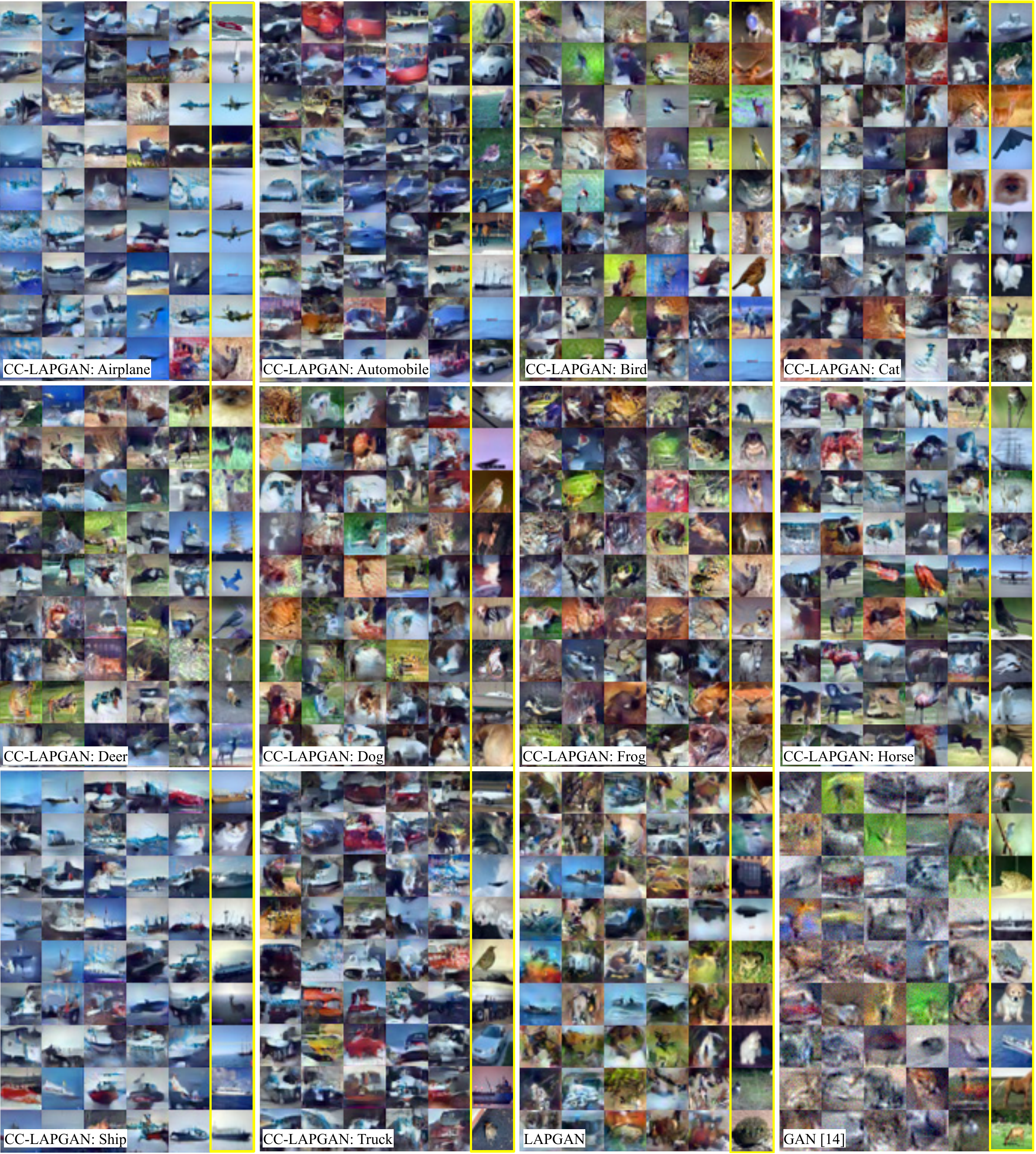} 
  \vspace{-2mm}
\caption{CIFAR10 samples: our class conditional CC-LAPGAN model,
  our LAPGAN model and the standard GAN model
of Goodfellow \cite{Goodfellow14}. The yellow column shows the
training set nearest
neighbors of the samples in the adjacent column. }
\label{fig:cifar_samples}
\end{figure}

%%%%%%%%%%%%%%%%%%%%%%%%%%%%%%%%%%%%%%%%%%%%%%%%%%%%%%%%%
%% STL10 samples 
%%%%%%%%%%%%%%%%%%%%%%%%%%%%%%%%%%%%%%%%%%%%%%%%%%%%%%%%%
% \begin{figure}[t]
% \centering
% \mbox{
% %\hspace{5mm}
% \subfigure[][]{
%   \includegraphics[width=0.5\linewidth]{imgs/stl10_64x64_sample.png} 
%   \label{fig:stl10_samples}
% }
% \hspace{1mm}
% \subfigure[][]{
%   \includegraphics[width=0.5\linewidth]{imgs/stl10_96x96_sample.png} 
%   \label{fig:stl10_samples}
% }
% }
% \caption{ {\bf(a)} Random 64x64 samples from the LAPGAN model trained on STL10 images. {\bf(b)} Random 96x96 samples from the LAPGAN model trained on STL10 images.}
% \label{samples}
% \end{figure}

%\begin{figure}[h!]
%\vspace{-2mm}
%\centering
%  \includegraphics[width=\linewidth]{imgs/stl10_crop2.pdf} 
%  \label{fig:stl_samples}
%\vspace{-5mm}
%\caption{Random 96x96 samples from the LAPGAN model trained on STL10 images.}
%\end{figure}

%%%%%%%%%%%%%%%%%%%%%%%%%%%%%%%%%%%%%%%%%%%%%%%%%%%%%%%%%
%% STL10 chain 
%%%%%%%%%%%%%%%%%%%%%%%%%%%%%%%%%%%%%%%%%%%%%%%%%%%%%%%%%
%\vspace{-5mm}

%%%%%%%%%%%%%%%%%%%%%%%%%%%%%%%%%%%%%%%%%%%%%%%%%%%%%%%%%
%% LSUN samples - regular and class conditional
%%%%%%%%%%%%%%%%%%%%%%%%%%%%%%%%%%%%%%%%%%%%%%%%%%%%%%%%%
% \begin{figure}[h!]
% %\vspace{-4mm}
% \centering
% \mbox{
%   \includegraphics[width=0.488\linewidth]{imgs/lsun_combo2.pdf}
%   \includegraphics[width=0.5\linewidth]{imgs/lsun_tower_64_vary_crop2.png}
%  }
%  \vspace{-5mm}
% \caption{LSUN samples at $64\times64$ from a class conditional LAPGAN
%   model church (upper left), bedroom (lower left) and  tower (right), seeded with validation set  $4\times4$ images (1st col). Each subsequent column shows draws from the model. }
% \label{fig:lsun_samples}
% \end{figure}

\subsection{Human Evaluation of Samples}
\begin{figure}[h!]
\vspace{-3mm}
\centering
\mbox{
        \includegraphics[width=0.35\textwidth]{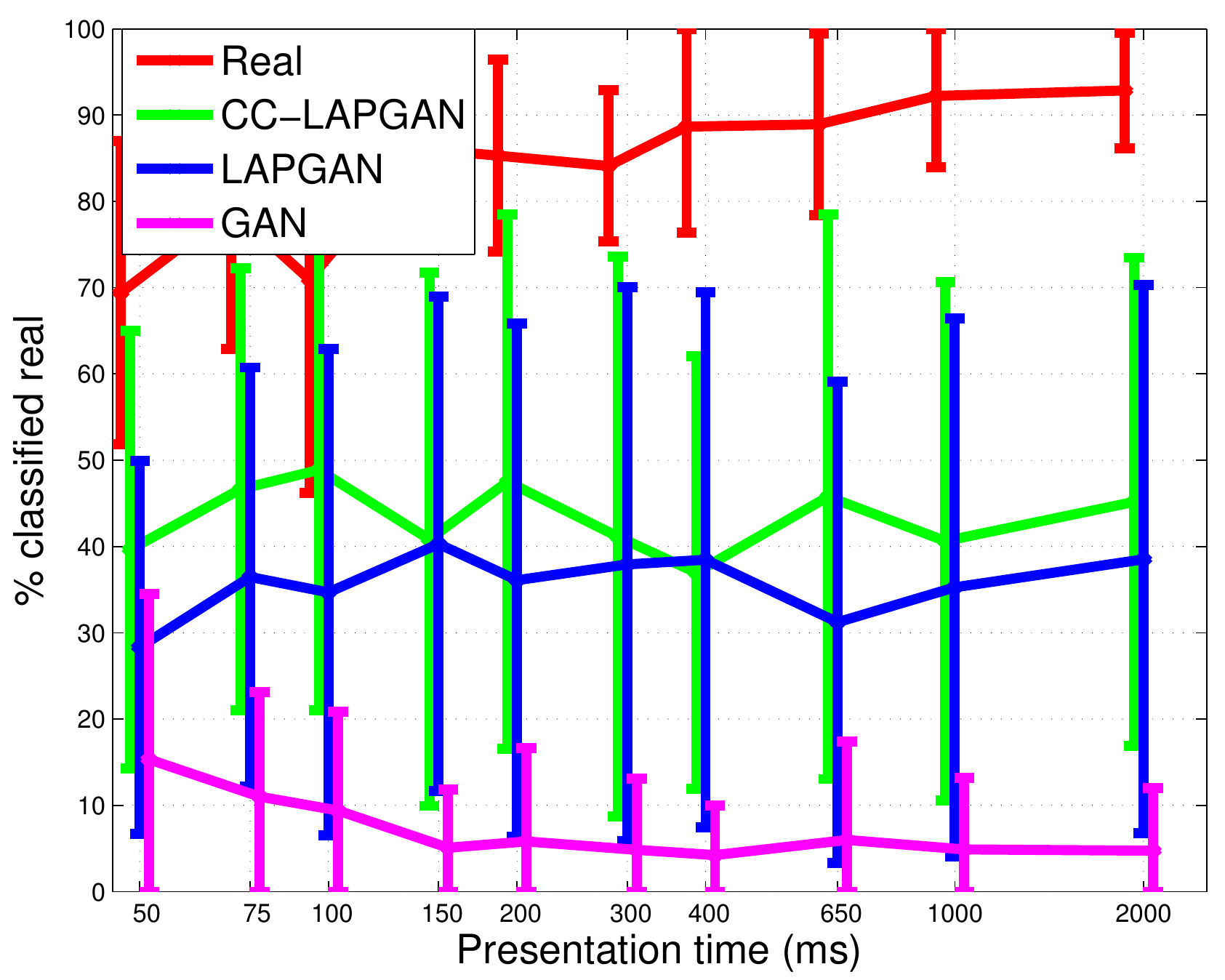}
\hspace{12mm}
       \fbox{ \includegraphics[width=0.25\textwidth]{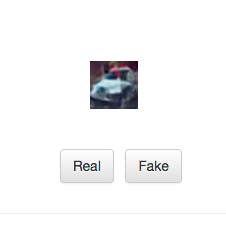}}
}
\vspace{-3mm}
\caption{Left: Human evaluation of real CIFAR10 images (red) and
  samples from Goodfellow \etal \cite{Goodfellow14} (magenta), our
  LAPGAN (blue) and a class conditional LAPGAN (green). The error
  bars show $\pm1\sigma$ of the inter-subject variability. Around 40\%
  of the samples generated by our class conditional LAPGAN model are realistic enough to fool a
human into thinking they are real images. This compares with 
$\leq10\%$ of images from the standard GAN model \cite{Goodfellow14}, but is
still a lot lower than the $>90\%$ rate for real images. Right:
The user-interface presented to the subjects.}
\label{fig:human}
\vspace{-3mm}
\end{figure}

To obtain a quantitative measure of quality of our samples, we asked
15 volunteers to participate in an experiment to see if they could
distinguish our samples from real images. The subjects
were presented with the user interface shown in \fig{human}(right) and
shown at random four different types of image: samples drawn from
three different GAN models trained on CIFAR10 ((i) LAPGAN, (ii) class
conditional LAPGAN and (iii) standard GAN \cite{Goodfellow14}) and also
real CIFAR10 images. After being presented with the image, the subject
clicked the appropriate button to indicate if they believed the image was
real or generated. Since accuracy is a function of viewing time, we
also randomly pick the presentation time from one of 11 durations
ranging from 50ms to 2000ms, after which a gray mask image is
displayed. Before the experiment commenced, they were shown examples
of real images from CIFAR10. After collecting $\sim$10k samples from the volunteers, we
plot in \fig{human} the fraction of images believed to be real for the four
different data sources, as a function of presentation time. The curves
show our models produce samples that are far more realistic than those
from standard GAN \cite{Goodfellow14}. 

